\documentclass{article}
\usepackage{booktabs}
\usepackage{amsmath}
\usepackage{amssymb}
\usepackage{multirow}
\usepackage{subfigure}
\usepackage{graphicx}
\usepackage{epstopdf}
\usepackage{color,soul}
\usepackage{algorithm}
\usepackage[noend]{algpseudocode}
\usepackage{rotating}
\usepackage{adjustbox}
\usepackage{enumerate}
\usepackage{lineno,hyperref}
\usepackage{pgfplots}
\usepackage{authblk}

\begin{document}

\title{A Fuzzy-Rough based Binary Shuffled Frog Leaping Algorithm for Feature Selection}

\author{Javad Rahimipour Anaraki\thanks{Memorial University of Newfoundland. Email:\texttt{jra066@mun.ca}}, Saeed Samet\thanks{University of Windsor. Email:\texttt{ssamet@uwindsor.ca}}, Mahdi Eftekhari\thanks{Shahid Bahonar University of Kerman. Email:\texttt{m.eftekhari@uk.ac.ir}}, \mbox{Chang Wook Ahn}\thanks{Gwangju Institute of Science and Technology. Email:\texttt{cwan@gist.ac.kr}}}

\date{}
\renewcommand\Affilfont{\itshape\small}
\providecommand{\keywords}[1]{\textbf{\textit{Keywords---}} #1}

\maketitle

\begin{abstract}
Feature selection and attribute reduction are crucial problems, and widely used techniques in the field of machine learning, data mining and pattern recognition to overcome the well-known phenomenon of the Curse of Dimensionality, by either selecting a subset of features or removing unrelated ones. This paper presents a new feature selection method that efficiently carries out attribute reduction, thereby selecting the most informative features of a dataset. It consists of two components: 1) a measure for feature subset evaluation, and 2) a search strategy. For the evaluation measure, we have employed the fuzzy-rough dependency degree (FRFDD) in the lower approximation-based fuzzy-rough feature selection (L-FRFS) due to its effectiveness in feature selection. As for the search strategy, a new version of a binary shuffled frog leaping algorithm is proposed (B-SFLA). The new feature selection method is obtained by hybridizing the B-SFLA with the FRDD. Non-parametric statistical tests are conducted to compare the proposed approach with several existing methods over twenty two datasets, including nine high dimensional and large ones, from the UCI repository. The experimental results demonstrate that the B-SFLA approach significantly outperforms other metaheuristic methods in terms of the number of selected features and the classification accuracy.
\end{abstract}


\section{Introduction}
Feature selection is the process of selecting the most informative features of a dataset while removing the others, and many feature selection methods have been in recent years \cite{Zhao,Hancer,Sreeja,Saha,Han,Lin,Canuto,Manimala}. The feature selection process results in a reduction in the size of datasets and a retention of their critical information. Features can be divided into three main groups: 1) irrelevant, 2) redundant, and 3) informative features. Irrelevant features are those which do not have any effect on the classification results, whereas redundant features are those which have a high correlation with other features except for the classification attribute(s). Finding and removing these two groups of features would reduce the size of datasets, thereby improving the classification accuracy as well as the visualization and comprehensibility of the induced concepts. The third group is the set of features that should remain at the end of the FS process.

Selecting $M$ out of $N$ features by means of a comprehensive search is an NP-hard problem \cite{Nock}. Furthermore, it has been proven that approximating the minimal relevant subset is hard up to very large factors \cite{Nock}. Therefore, greedy search methods and metaheuristic search strategies are suitable for solving this problem \cite{Yusta}. However, all of the greedy search methods suffer from the deficiency of becoming trapped in local optima \cite{Yusta}. Forward and backward search mechanisms are instances of greedy search algorithms that are widely used for FS, because of their ideal time complexity; therefore, they are not capable of avoiding local optima \cite{Yusta,Pudil}. Due to this deficiency and the inherent ability of metaheuristic search methods to find the global optimum while avoiding local optima, these search methods have been widely utilized to solve FS problems \cite{Yusta,Elalami,Nemati,Vieira}.

Genetic algorithm (GA), particle swarm optimization (PSO), Tabu search and memetic algorithms are representative metaheuristic instances that, in recent years, have been very successful at solving various NP-hard engineering problems such as feature selection \cite{Yusta,Elalami,Nemati,Vieira}. 
Moreover, all of the above search mechanisms require an evaluation criterion for measuring the suitability of feature subsets. Based on determining the evaluation measures, a twofold taxonomy of feature selection methods has been presented in the literature \cite{Sebban}. In this taxonomy, feature selection strategies are categorized into 1) filter-based methods, and 2) wrapper-based methods. The former generally evaluate a feature subset by performing statistical tests on the data \cite{Sebban}. Thus, the filter-based methods ``filter out'' irrelevant features before the induction process (i.e. classification). In the wrapper-based approach, an induction algorithm itself (i.e. classifier) is utilized for evaluating feature subsets \cite{Sebban}. In other words, it is used for optimizing the accuracy rate estimated by an induction algorithm. Compared to filter-based methods, wrapper-based methods are computationally prohibitive since they employ an induction model as an embedded algorithm. On the other hand, the wrapper-based methods are more accurate at finding a proper subset of informative features than filter-based methods. In the filter-based technique, a non-statistical criterion can also be used as the evaluation measure. Examples of such criteria include the dependency degree (DD) based on rough set theory \cite{Thangavel}, and the fuzzy feature saliency measure \cite{Verikas} based on fuzzy set theory.
Recently, much research has been performed on the development of methodologies for dealing with imprecision and uncertainty \cite{Thangavel,Verikas,Degang}. Fuzzy and rough set theories are analogous in the sense that they can model uncertainty and inconsistency. Recent studies have shown that they are complementary in nature. 

Fuzzy-rough feature selection (FRFS) is one of the most successful hybrid tools for dimensional reduction, which is capable of handling both discrete and real-valued (or a mixture of both) variables \cite{Degang}. However, there are some problems regarding the use of FRFS, thoroughly addressed in \cite{Jensen}. For instance, pre-data discretization by using fuzzy partitions is an FRFS approach that is not very successful in terms of computation. One of the newly developed FRFS methods is the lower approximation-based fuzzy-rough feature selection (L-FRFS) \cite{Jensen} method. L-FRFS, introduced in \cite{Jensen} is a fast FRFS, and it exhibits better performance compared to previously developed FRFSs. Moreover, as stated earlier, generating all subsets of features is an NP-hard problem and computationally prohibitive. Therefore, some hill-climbing search algorithms have been proposed in the literature in order to compensate for this computational deficiency \cite{Jensen}. 

The smallest subset of features with the highest DD is called the ``minimal reduct''; it might not be found by the fuzzy-rough QuickReduct algorithm, which is an example of a hill-climbing method, both in terms of the resulting dependency measure and the subset size. Due to the deficiencies of hill-climbing approaches, metaheuristic algorithms such as GA and PSO are required in order to find such minimal reducts, especially when available data are high-dimensional. In \cite{Chen,Suguna,Wang,wroblewski,Anaraki2}  metaheuristic algorithms and rough set theory have been combined to find minimal reducts. In recent years, a few studies have also been presented in literature regarding the hybridization of fuzzy-rough and metaheuristic approaches \cite{Degang,Jensen}. Very significant work is the combination of ACO and fuzzy-rough set for dimension reduction \cite{Jensen2}. In this work, Jensen and Shen utilized a computationally demanding FRFS method in which continuous data have been discretized in advance by fuzzy partitions, and an ACO has been employed to find the minimal reduct \cite{Jensen2}. As mentioned earlier, the authors have recently confirmed the time deficiencies of the fuzzy-rough method used in \cite{Jensen}, and as an alternative have introduced the L-FRFS as a fast method. 

In \cite{Xiang}, Xiang et al. have proposed a hybrid method for feature selection by improving the diversity of species through piecewise linear chaotic maps (PWL), and increasing the speed of local search by applying sequential quadratic programming (SQP) to the binary gravitational search algorithm (GSA). The improved version of GSA has been hybridized with a $1$-nearest neighbour method to from a feature selection system. A modified version of the binary PSO with the ability to avoid premature convergence utilizing both velocity and similarity of best solutions has been introduced by Vieira et al. \cite{Vieira1}. The search method has been used to perform simultaneous feature selection and prediction of mortality of septic patients using concurrently optimized kernel parameters of a support vector machine (SVM). On of the most recent and successful feature selection methods is gradient boosted feature selection (GBFS) proposed by Xu et al. \cite{Xu}. It works based on gradient boosted trees \cite{Friedman2001}. It starts by building regression trees using CART algorithm \cite{Breiman}, and features are selected simultaneously based on deviation in impurity function. Selecting new feature is penalized and reusing already selected features has no cost. 

In the present paper, a new FRFS technique is proposed on the basis of the B-SFLA and L-FRFS. Our contributions are twofold: 1) we devise a new binary version of an SFLA that employs a new dissimilarity measure, new coefficients for self-parameter selection, and a modified ranking rule, and 2) we develop an FS method by combining the strengths of this B-SFLA and the L-FRFS. 
The rest of this paper is organized as follows. Section \ref{motivation} shows our motivation. In Section \ref{back}, the background of fuzzy-rough feature selection and the search algorithm are presented. Section \ref{pro} illustrates the proposed feature selection method. Section \ref{res} reports experimental results and finally we conclude this paper in Section \ref{con}.

\section{Motivation} \label{motivation}
Fuzzy-rough feature selection (FRFS) works based on fuzzy-rough dependency degree (FRDD). This measure varies from zero to one, which reflect not dependent to totally dependent of outcome to features, respectively. Moreover, having several subsets with the highest dependency degree is desirable in the search process.
Based on literature, previous feature selection methods using FRFS and variety of search algorithms such as GA, and PSO were not be able to find more than one subset even for small datasets. This incapability motivated us to modify, and hybridize binary shuffled frog leaping algorithm (B-SFLA) with FRDD toward finding the smallest possible subsets of features with the greatest possible value for FRDD. To this end, we have modified underlying components of B-SFLA such as distance measure to meet our goal.

\section{Background}\label{back}
\subsection{Fuzzy-Rough Feature Selection} \label{frf}
Rough set theory was proposed by Pawlak as a tool to deal, in an efficient way, with uncertainty \cite{Pawlak}, in data organized in a decision table. Let $U$ be the universe of discourse and $A$ be a nonempty finite set of attributes in $U$; information system is shown by $I=(U,A)$. Let $X$ be a subset of $U$, and $P$ and $Q$ be two subsets of $A$; approximating a subset using rough set theory is done by means of upper and lower approximations. The upper approximation of $X$ with regard to $(\overline{P}X)$ contains objects, which are possibly classified in $X$ regarding the attributes in $P$. Objects in the lower approximation $(\underline{P}X)$ are those, which are definitely classified in $X$ regarding the attributes in $P$. A rough set is shown by an ordered pair, $(\underline{P}X, \overline{P}X)$. Different regions are defined using this pair as shown in Eqs. \ref{pos} to \ref{bnd}.
\begin{equation}\label{pos}
POS_{P}(Q) = \bigcup_{X \in \mathbb{U}/Q} \underline{P}X
\end{equation}

The positive region of partition $\mathbb{U}/Q$ is a set of all objects, which can be uniquely classified into blocks of the partition by means of $P$.
\begin{equation}\label{neg}
NEG_{P}(Q) = \mathbb{U} - \bigcup_{X \in \mathbb{U}/Q} \overline{P}X
\end{equation}

The negative region is a set of objects, which cannot be classified into the partition $\mathbb{U}/Q$, \cite{komorowski}. 
\begin{equation}\label{bnd}
BND_{P}(Q) = \bigcup_{X \in \mathbb{U}/Q} \overline{P}X - \bigcup_{X \in \mathbb{U}/Q} \underline{P}X
\end{equation}

The boundary region of $X$ can be determined by subtracting the lower approximation from the upper approximation; if it is a non-empty set, $X$ is called a rough set, otherwise, is a crisp set \cite{Pal}.

Finding the dependency between attributes is one of the most important areas in data analysis. The dependency of $Q$ on $P$ is denoted by $P\Rightarrow_k Q$ and $k = \gamma_{p}(Q)$, in which $\gamma$ is the dependency degree \cite{komorowski}. If $k=1$ then $Q$ completely depends on $P$ and if $0<k<1$ then $Q$ partially depends on $P$.
The value of $k$ is a measure of the dependency between the features $P$ and $Q$. In feature selection, features which have lower dependency on each other and are highly correlated to the decision feature(s), are desired. If $Q$ completely depends on $P$, then the partition which is made by $P$ is finer than $Q$. The positive region of the partition $\mathbb{U}/Q$, with respect to $P$, which is denoted by $POS_{P}(Q)$, is the set of all elements which can be classified into the partition $\mathbb{U}/Q$ using $P$ \cite{komorowski}. The following equation allows to calculate the dependency.
\begin{equation}\label{dd}
k = \gamma_{P}(Q) = \frac{|POS_{P}(Q)|} {|\mathbb{U}|},
\end{equation}
where notation $| . |$ is used for cardinality. The reduct is a subset of features which have the same dependency degree as employing all the features for classification. The features that belong to the reduct set are the most informative ones while the others are either irrelevant or redundant.

In many cases, we face datasets containing crisp and real-valued data together and a rough set cannot handle real-valued data. The need for a method based on rough set theory for resolving this problem has been a motivation for researchers to combine fuzzy theory with rough set theory. Rough set and fuzzy set theories have a connection; they both handle the information granulation problem. Fuzzy set theory focuses on fuzzy information granulation, but rough set theory concentrates on crisp information granulation \cite{Degang,Jensen,Radzikowska}. 
One way to handle real-valued data using rough set theory is to discretize continuous data in advance and make a new crisp valued dataset. Discretization is not enough as long as the similarity between two values remains unspecified \cite{Jensen}. Therefore, dependency degree between the features is calculated by means of the FRDD. The fuzzy-rough set basis will be addressed thoroughly in Section \ref{pro}.

In this paper, three search algorithms -- namely, genetic algorithm (GA), particle swarm optimization (PSO) and shuffled frog leaping algorithm (SFLA) -- are investigated to resolve the problem of finding the minimal reduct.

\subsection{Genetic Algorithm}
Genetic algorithms (GA), proposed by Holland \cite{holland}, are an iterative heuristic search algorithms that use natural evolution to solve optimization problems. The search begins with a random population conditioned by genomes. Each genome consists of a bit string. Two genomes are combined using a \emph{crossover} operator based on their fitness values. The \emph{crossover} operation can be done at one or more points. The population of genomes is called parent population and the results from their combination are called the offspring. The fitness value of a genome is calculated using a \emph{fitness function}. Another operator which changes the value of a genome is called a \emph{mutation}. This operator maintains population diversity and is used for avoiding local optima. For \emph{mutation}, a randomly selected bit in a genome changes to its complementary state. After each generation, the best individuals from the offspring are selected according to a \emph{selection} mechanism to be moved to the next generation of parents. 

\subsection{Particle Swarm Optimization}
Particle swarm optimization (PSO), proposed by Kennedy and Eberhard \cite{Kennedy}, is an evolutionary search algorithm which was inspired by the behaviour of natural swarms such as fish schools and birds flocks. In this algorithm, a solution is encoded as a string which is called a \emph{particle}. All particles have a fitness which is calculated using a \emph{fitness function}. Each particle moves over the search space based on its current position, velocity, its own experience, and the best particle experience. In \cite{Gonzalo} the authors review PSO and its modifications, and maintain that performing a stochastic stability analysis of the particle trajectories is vital to convergence.

\subsection{Shuffled Frog Leaping Algorithm}
The Shuffled frog leaping algorithm (SFLA) is a memetic metaheuristic search algorithm proposed by Eusuff et al. \cite{Eusuff}; it is basically a combination of a shuffled complex evolution (SCE) algorithm \cite{Duan} that ensures global exploration, and PSO \cite{Kennedy} that is responsible for local search. Randomness and determinism are the results of this combination. The SFLA is based on memetics of frog-like beings. A meme is an idea or information pattern which is replicated or repeated to someone else. Memes and genes are analogous but are different in the way they propagate. A \emph{meme} is propagated by leaping from one brain to another and can be transmitted between any individual, but a gene is propagated from parent to offspring by (sexual) reproduction. 

The algorithm is inspired by real frog populations searching for food. In this algorithm, the behaviour of the population is determined by memes, and thus the population is more important than individuals. In the SFLA, frogs are partitioned into memeplexes that are evaluated individually. In each memeplex, frogs are influenced by each other and they experience meme evolution. Memetic evolution increases the frogs' performance in terms of reaching the goal by using information from the memeplex and the best performing individual in the population. This process continues for a predefined number of iterations. Then, all memeplexes are mixed with each other to form a new set of memeplexes through shuffling. Frogs with better performance contribute more to distribute new individuals in the population. A modified version of the SFLA has been proposed by Reddy et al. \cite{Reddy} for solving the environmentally-constrained economic dispatch problem. The modified algorithm uses a local search as well as a new parameter to accelerate convergence.

\section{Proposed Feature Selection Approach} \label{pro}
In this section, the proposed approach is defined based on the two main concepts of feature selection: 1-evaluation measure, and 2- search method. The evaluation measure is fuzzy-rough dependency degree (FRDD) and the search method is a binary modification of SFLA.

\subsection{Evaluation Measure} \label{eval}
The QuickReduct algorithm finds a reduct set without finding all the subsets\cite{Jensen}. It begins with an empty set and each time selects the feature that causes the greatest increase in dependency degree (DD). The algorithm stops when adding more features does not increase the DD. Since it employs a greedy algorithm, it does not guarantee that the minimal reduct set will be found. For this reason, a new FRFS algorithm is presented in this paper. Prior to providing the details of our approach, it is necessary to introduce the definition of the FRDD. To begin with, the definition of the $X$-lower and $X$-upper approximations and the degree of fuzzy similarity \cite{Jensen} are given by Eqs. \ref{lowerJ} to \ref{combin}, respectively.
\begin{equation}\label{lowerJ}
\mu_{\underline{R_P}X} (x) =\inf_{y \in \mathbb{U}} I\{\eta_{R_P}(x, y), \mu_{X}(y)\},
\end{equation}

\begin{equation}\label{upperJ}
\mu_{\overline{R_P}X} (x) =\sup_{y \in \mathbb{U}} T\{\eta_{R_P}(x, y), \mu_{X}(y)\},
\end{equation}

\begin{equation}\label{combin}
\eta_{R_P}(x, y) = \bigcap_{a \in P} \{\eta_{R_a}(x, y)\},
\end{equation}
where $I$ is a \L ukasiewicz fuzzy \emph{implicator}, which is defined by $min(1-x+y,1)$, and $T$ is a \L ukasiewicz fuzzy \emph{t}-norm, which is defined by \mbox{$max(x+y-1,0)$}. In \cite{Radzikowska}, three classes of fuzzy-rough sets based on three different classes of implicators, namely \emph{S}-, \emph{R}-, and \emph{QL}-implicators, and their properties have been investigated. Here, $R_P$ is the fuzzy similarity relation considering the set of features in $P$, and $\eta_{R_P}(x, y)$ is the degree of similarity between objects $x$ and $y$ over all features in $P$. Also, $\mu_{X}(y)$ is the membership degree of $y$ to $X$. One of the best fuzzy similarity relations as suggested in\cite{Jensen} is given by Eq. \ref{sim}.
\begin{equation}\label{sim}
\eta_{R_a}(x, y) = max\left\{min\left\{\frac{(a(y)-(a(x)-\sigma_a))}{(a(x)-(a(x)-\sigma_a))}, \frac{((a(x)+\sigma_a)-a(y))}{((a(x)+\sigma_a)-a(x))}\right\}, 0\right\},
\end{equation}
where $\sigma_a$ is variance of feature $a$. The L-FRFS does not use the fuzzy partitioning used in FRFS, and thereby it is more computationally effective.

The FRFS can be conducted on the real-valued datasets using the lower approximation. The positive region in rough set theory is defined as a union of lower approximations. Referring to the extension principle \cite{Jensen}, the membership of object \emph{x} to a fuzzy positive region is given by Eq. \ref{posN}.
\begin{equation}\label{posN}
\mu_{POS_{P}(Q)} (x) =\sup_{X \in \mathbb{U}/Q} \mu_{\underline{P}X}(x).
\end{equation}

If the equivalence class that includes $x$ does not belong to a positive region, clearly \emph{x} will not be part of a positive region. Using the definition of positive region, the FRDD function \cite{Jensen} is defined as:

\begin{equation}\label{fdd}
\gamma'_{P}(Q) = \frac{|\mu_{POS_{P}(Q)} (x)|} {|\mathbb{U}|}=\frac{\sum_{x \in \mathbb{U}} \mu_{POS_{P}(Q)} (x)}{|\mathbb{U}|}.
\end{equation}

Based on the concept of the FRDD, we have developed a new metaheuristic search mechanism in order to effectively discover the minimal reducts. Among various search algorithms, such as GA and PSO, the SFLA can be used as a promising search method for feature selection (which is an NP-hard problem), due to its performance toward global optimal solution, both from a likelihood and a speed perspective \cite{Eusuff}. Based on the published results in \cite{Eusuff}, the GA has failed to find best values in 20\% of the cases, and it also needs a higher number of function evaluations to find the optimal value, compared to the SFLA. The SFLA is capable of finding a subset of solutions along with the optimal answer as the final result. Since the feature selection problem is fundamentally binary, the need for a binary search algorithm is inevitable. 

\subsection{Search Method} \label{search}
The search process starts by randomly initializing each binary individual with the size of the number of features, and continues by participating in ranking, partitioning and evolutionary processes. Generally, the SFLA consists of seven steps as follows:

\begin{enumerate}[Step 1]

\item\textbf{\textit{\underline{Initialize the population:}}} Choose $m$ and $n$. Here, $m$ is the number of memeplexes, and $n$ is the number of frogs in each memeplex. The total number of frogs is then $F=m \times n$.

\item\textbf{\textit{\underline{Generate a population:}}} The total number of frogs in the feasible space is \mbox{$\Omega \subset \mathbb{\Re}^{d}$} where $d$ is the number of decision variables (features); the \emph{i}th frog is encoded as \mbox{$U(i)=(U_i^1, U_i^2, ..., U_i^d)$}. Compute the fitness value for all individuals using Eq. \ref{fdd}.

\item\textbf{\textit{\underline{Rank frogs:}}} Sort frogs in descending order of their fitnesses, and record them in $X=\{U(i), f(i), i=1, ..., F\}$. The position of the first (i.e., best) frog is recorded in $P_X$, where $P_X=U(1)$ .

\item\textbf{\textit{\underline{Partition frogs into memeplexes:}}} Partition the array $X$ of frogs into $m$ memeplexes, each containing $n$ frogs.
\begin{equation} \label{rankfrog}
\begin{split}
Y^k=&[U{(j)}^k, f{(j)}^k | U{(j)}^k = U(k+m(j-1)), \\f{(j)}^k=&f(k+m(j-1)), j=1, ..., n], k=1, ..., m
\end{split}
\end{equation}

\item\textbf{\textit{\underline{Memetic evolution in each memeplex:}}} Each memeplex is involved in the evolution which is described later in the Step 5's subsection.

\item\textbf{\textit{\underline{Shuffle memeplexes:}}} After a predefined number of evolution rounds, all memeplexes are mixed into $X$, and sorted in descending order.

\item\textbf{\textit{\underline{Check convergence:}}} If the convergence criteria are satisfied, stop. Otherwise, go to Step 4.
\end{enumerate}

Note that in the Step 5, the evolution process is repeated $N$ times. This process is comprised of further steps, as follows:

\begin{enumerate}[Step 5.1]
\item\textbf{\textit{\underline{Initialization:}}} Set $i_m=0$ and $i_N=0$ as two counters for memeplexes and evolutions, respectively.

\item $i_m=i_m+1$

\item$i_N=i_N+1$

\item\textbf{\textit{\underline{Construct a submemeplex:}}} In order to avoid being trapped in local optima, a subset of memeplexes is selected for moving toward. The submemeplex selection strategy is based on a triangular probability distribution (see Eq. \ref{submem}) that assigns the highest value to a frog with the maximum fitness and the lowest value to a frog with the minimum fitness. This assignment increases the chances of a high performing frog being selected.
\begin{equation}\label{submem}
p_j=\frac{2 \times (n+1-j)}{n \times (n+1)}, j=1,...,n
\end{equation}
For example, for $j = 1$ and $j = n$, the probabilities are given by:
\begin{equation*}
p_1=\frac{2}{n+1}, p_n=\frac{2}{n \times (n+1)}
\end{equation*}
After the submemeplex formation, it is sorted in descending order in an array, $Z$, and the best and the worst positions are recorded in $P_B$ and $P_W$, respectively.

\item\textbf{\textit{\underline{Improve the worst frog:}}} The worst frog's position is improved using Eqs. \ref{posStep} and \ref{negStep} for positive and negative steps, respectively.
\begin{equation}\label{posStep}
\textrm{step size }S = min\{int\{rand \times (P_B - P_W)\}, S_{max}\}
\end{equation}
\begin{equation}\label{negStep}
\textrm{step size }S = max\{int\{rand \times (P_B - P_W)\}, -S_{max}\},
\end{equation}
where \emph{rand} is a random number, \emph{int} is the integer part of a number, and $S_{max}$ is the maximum step size allowed to be adopted after infection. Since the $P_B$ and $P_W$ are in binary form, the distance between two parameters is calculated using the \emph{HD}; therefore, Eqs. \ref{posStep} and \ref{negStep} are modified to Eqs. \ref{newposStep} and \ref{newnegStep} to deal with binary parameters.
\begin{equation}\label{newposStep}
\textrm{step size }S = min\{int\{rand \times HD(P_B, P_W)\}, S_{max}\}
\end{equation}
\begin{equation}\label{newnegStep}
\textrm{step size }S = max\{int\{rand \times HD(P_B, P_W)\}, -S_{max}\}.
\end{equation}

Then, the new position is calculated by:
\begin{equation}\label{newPos}
U_{(q)}=P_W+S,
\end{equation}
where $q$ is the number of randomly selected frogs from $n$ frogs to form a memeplex and it is initialized manually. If $U_{(q)}$ is in feasible space $\Omega$, then compute the fitness value, $f_{(q)}$; otherwise, go to the Step 5.6. If the newly computed $f_{(q)}$ is better than the old $f_{(q)}$, then go to the Step 5.8; otherwise, go to the Step 5.6. 

In the binary form, a vector of random values of size $S$ is generated. Then for each \emph{HD}, a random number is produced; if it is greater than the specified random value in the vector, then that bit will remain unchanged, otherwise, it will be changed to the corresponding bit of the best frog.
For example, the \emph{HD} of $P_B=11001010$ and $P_W=10101000$ is 3. Then a random vector with three elements such as $\{0.11, 0.05, 0.96\}$ along with three random numbers for each position, such as $Pos_2 = 0.74$, $Pos_3 = 0.60$, and $Pos_7 = 0.79$ are produced. Finally, by comparing each position's random number with the same position's random value in the random vector, the must-change bit(s) will be specified. In this example, the seventh bit must be changed since its related random value in the vector is higher than its position's random number; thus, the final result would be $P_W=10101010$. 

\item Compute a new position using Eqs. \ref{posStepX} and \ref{negStepX}. The binary forms are shown in Eqs. \ref{newposStepX} and \ref{newnegStepX}. 
\begin{equation}\label{posStepX}
\textrm{step size }S = min\{int\{rand \times (P_X - P_W)\}, S_{max}\}
\end{equation}

\begin{equation}\label{negStepX}
\textrm{step size }S = max\{int\{rand \times (P_X - P_W)\}, -S_{max}\}
\end{equation}

\begin{equation}\label{newposStepX}
\textrm{step size }S = min\{int\{rand \times HD(P_X, P_W)\}, S_{max}\}
\end{equation}

\begin{equation}\label{newnegStepX}
\textrm{step size }S = max\{int\{rand \times HD(P_X, P_W)\}, -S_{max}\}.
\end{equation}
If $U_{(q)}$ is in feasible space $\Omega$, then compute the fitness value, $f_{(q)}$; otherwise, go to Step 5.7. If the newly computed $f_{(q)}$ is better than the old $f_{(q)}$, then go to Step 5.8; otherwise, go to Step 5.7.

\item\textbf{\textit{\underline{Censorship:}}} Replace this frog with a randomly generated frog, $r$. 

\item\textbf{\textit{\underline{Update the memeplex:}}} After changing the worst frog's position in the submemeplex, replace $Z$ in their original locations in $Y^{i_m}$ . Sort $Y^{i_m}$ in descending order. 

\item If $i_N<N$ , go to Step 5.3.

\item If $i_m<m$ , go to Step 5.2.
\end{enumerate}

Meanwhile, a modification for calculating the distance of the frogs is further applied to the proposed binary SFLA. The distance of the frogs that was calculated using the \emph{HD} is replaced with a dissimilarity measure based on the fuzzy-rough set. The positive region i.e., $POS(.)$ \cite{Kamyab} as presented in Eq. \ref{posN} is used instead of the \emph{HD}. The positive region sees the frogs as features and calculates the similarity between each frog and the best frog. The value of $POS(.)$ varies from zero to the number of the variables. Since this distance must be dissimilarity, this measure is subtracted from the length of the binary frog. 
This measure can be employed in the Step 5, and the modified equations are given by Eqs. \ref{modposStepX} and \ref{modnegStepX} are used in the Step 5.6. 
\begin{equation}\label{modposStepX}
\textrm{step size }S = min\{int\{rand \times (L-POS(P_B, P_W))\}, S_{max}\}
\end{equation}

\begin{equation}\label{modnegStepX}
\textrm{step size }S = max\{int\{rand \times (L-POS(P_B, P_W))\}, -S_{max}\},
\end{equation}
where $L$ is the length of a binary frog, and $S_{max}$ is the maximum step size allowed to be adopted after evolution. 

The hybridization of the B-SFLA with FRDD is suggested to discover more than one reduct with the highest dependency degree. The L-FRFS can be considered as a multi-modal problem, in which the smallest subset of features with the highest FRDD is desired. Thus, conventional evolutionary algorithms might find many global optima with the highest FRDD; however, a question arising here is ``which one is the best?''; Referring to the fitness, all of these solutions are acceptable, whereas referring to the cardinality of the subsets they varies. By ranking the subsets with the same FRDD, based on the number of selected features, a new wide range of reducted subsets is provided. This range can be analyzed using the frequency of a feature's appearance in all of the reducted subsets. The most frequent features might play an important role in specifying the outcome. 

The aforementioned strategy is placed in the Step 5.4 of meme evolution and the Step 3, ranking frogs, of the B-SFLA; however, the ranking process is primarily based on the FRDD and in the case of having several subsets with the identical FRDD, it ranks subsets based on their cardinality. Through this process, the B-SFLA returns more than one reduct in a single run; conventional search methods do not always return more than one reduct. These minimal sets satisfy both criteria: the highest FRDD and the lowest number of selected features.

Using this method, the frogs leap toward two goals simultaneously. In the very first leaps, frogs jump toward the subsets with the highest FRDD; therefore, they try to increase their fitness as much as possible. In the following leaps, when the number of frogs with the maximum fitness is increased, the population selects the individuals with both the highest FRDD and the lowest number of features. Algorithm \ref{algorithm} shows pseudo code of the proposed method.

\begin{algorithm}[!h]
\caption{FRFS based on B-SFLA}
\label{algorithm}
\begin{algorithmic}[1]
\Procedure{search\textendash evaluate}{}
\State initialize $m, n, q, N, S_{max}$
\State generate a population of $(m \times n)$ frogs
\State rank frogs in $X$ based on the number of features and FRDD
\State partition $X$ into $m$ memeplexes $Y^1, Y^2, ..., Y^m$
\While{$i_m<m$}
\While{$i_N<N$}
\State construct a submemeplex containing $q$ frogs into $Z$
\State improve the worst frog in $S_{max}$ steps, update FRDD
\State replace infeasible and halting frogs
\State partition $Z$ into $Y^1, Y^2, ..., Y^m$
\EndWhile
\State \textbf{end}
\EndWhile
\State \textbf{end}
\State combine $Y^1, Y^2, ..., Y^m$ into $X$ and update the best frog
\State check the convergence criteria
\EndProcedure
\State \textbf{end}
\end{algorithmic}
\end{algorithm}

In the preparation section, parameters of the B-SFLA are initialized based on the properties of the current dataset. Then, $m \times n$ diverse subsets of features are evaluated and evolved based on FRDD and B-SFLA, respectively. Then, the outcome of the algorithm is fed to nine different classifiers to avoid any tendency toward specific classification method. Finally, the mean of the resulting classification accuracies is calculated.

Since the complexity of meta-heuristic search algorithms are very depended on their parameters, it is worth mentioning that the complexity of the FRDD is $O(n^2)$ in the worst case \cite{JensenComputational}, where $n$ is number of features.

Our modified version of B-SFLA has the following advantages toward selecting high quality features:
\begin{enumerate}
\item It finds multiple subsets of features as the final result
\item It uses FRPR as a distance measure to distinguish dissimilarity of individuals
\item Automatic and optimized parameter selection based on the number of features in each dataset
\end{enumerate}

\section{Experimental Results and Discussion} \label{res}
Twenty two datasets from the UCI repository of machine learning \cite{UCI} including nine large datasets -- namely, LSVT Voice Rehabilitation \cite{LSVT}, Urban Land Cover \cite{UrbanLandCover1,UrbanLandCover2}, Arrhythmia, Molecular Biology, COIL 2000 \cite{COIL}, CNAE-9, Madelon \cite{Guyon}, MicroMass, and Arcene \cite{Guyon} -- have been selected and used to perform a comparative study. These datasets and their characteristics are shown in Table \ref{spec}. The table is sorted based on the number of samples $\times$ features.

\begin{table}[h]
\caption{Dataset characteristics}
\centering
\begin{tabular}{c c c}
\hline
Datasets & Samples & Features\\
\hline
Breast Tissue&106&10\\
Lung Cancer&32&56\\
Glass&214&10\\
Wine&178&13\\
Olitos&120&25\\
Heart&270&13\\
Cleveland&303&13\\
Parkinson&197&23\\
Pima Indian Diabetes&768&8\\
Breast Cancer Wisconsin&699&10\\
Ionosphere&351&33\\
Sonar&208&60\\
Libras Movement&360&90\\
LSVT Voice Rehab.&126&310\\
Urban Land Cover&675&148\\
Arrhythmia&452&279\\
Molecular Biology&3190&60\\
COIL 2000&5822&85\\
CNAE-9&1080&857\\
Madelon&2000&500\\
MicroMass&931&1300\\
Arcene&200&10000\\
\hline
\end{tabular}
\label{spec}
\end{table}

The \emph{fitness function} for all of the search algorithms is the FRDD depicted in Eq. \ref{fdd}. The GA and PSO parameters are presented in Tables \ref{GA} and \ref{PSO}, respectively. For both algorithms, the population size and the number of generations are identical to B-SFLA's to enable further comparisons.
As presented in \cite{Eusuff}, the SFLA parameter selection should be performed based on the properties of the problem. Parameter selection is one of the most important aspects of using search algorithms; however, it is still untouched for feature selection. Referring to the authors' recommendation in \cite{Eusuff}, for problems with 15-20 variables, the ranges in Table \ref{SFLA} are suggested. However, the parameter selection for feature selection has been formulated based on the total number of all features ($all\_F$) using a trial and error method. The results are shown in Table \ref{proSFLA}. 
Further investigations show that the proposed parameters in Table \ref{proSFLA} work remarkably well for small datasets with less than 15,000 data cells; however, parameters in Table \ref{proSFLAall} \cite{Eusuff} can be used not only for small and medium datasets, but also for large ones. 

\begin{table}[h]
\caption{GA parameters}
\centering
\begin{tabular}{c c c c}
\hline
Population & Generation & $P_c$ & $P_m$\\
\hline
900&5&0.600&0.033\\
\hline
\end{tabular}
\label{GA}
\end{table}

\begin{table}[h]
\caption{PSO parameters}
\centering
\begin{tabular}{c c c c}
\hline
Particles & Iteration & {$C_1$} & {$C_2$}\\
\hline
900&5&2&2\\
\hline
\end{tabular}
\label{PSO}
\end{table}

\begin{table}[!h]
\caption{SFLA parameters}
\centering
\setlength{\tabcolsep}{3.8pt}
\begin{tabular}{c c c c c}
\hline
\bf{$m$}&\bf{$n$}&\bf{$N$}&\bf{$q$}&\bf{$S_{max}$}\\
\hline
$100 \leq m \leq 150$ & $30 \leq n \leq 100$ & $20 \leq N \leq 30$ & 20 & $1.00 \times all\_F$\\
\hline
\end{tabular}
\label{SFLA}
\end{table}

\begin{table}[!h]
\caption{Proposed B-SFLA parameters for datasets with size of data cells $\le 15,000$}
\centering
\setlength{\tabcolsep}{2.5pt}
\begin{tabular}{*{5}{c}}
\hline
\bf{$m$}&\bf{$n$}&\bf{$N$}&\bf{$q$}&\bf{$S_{max}$}\\
\hline
$2.20 \times all\_F$ & $0.70 \times all\_F$ & $0.50 \times all\_F$ & $0.45 \times all\_F$ & $0.50 \times all\_F$\\
\hline
\end{tabular}
\label{proSFLA}
\end{table}

\begin{table}[!h]
\caption{Proposed B-SFLA parameters for most datasets}
\centering
\begin{tabular}{c c c c c}
\hline
\bf{$m$}&\bf{$n$}&\bf{$N$}&\bf{$q$}&\bf{$S_{max}$}\\
\hline
30 & 30 & 5  & 15  & $0.45 \times all\_F$\\
\hline
\end{tabular}
\label{proSFLAall}
\end{table}

The number of selected features obtained by each search algorithm is shown in Table \ref{selF}. In terms of the number of selected features, the GBFS has selected the least number of features compared to the other methods; however,  selecting one feature as a final result for Breast Tissue, Lung Cancer, Glass, Wine, and Sonar is not desirable both from an in-field and a data processing point of view. Selecting a very small number of features reduces the utility of feature selection methods for pre-processing and model complexity improvement.

\begin{table}[h]
\caption{Number of selected features obtained by each search algorithm}
\centering
\setlength{\tabcolsep}{3.7pt}
\begin{tabular}{c c c c c c}
\hline
Datasets & L-FRFS & GA & PSO & GBFS &B-SFLA\\
\hline
Breast Tissue&9&9&9&1&4\\
Lung Cancer&6&7&4&1&3\\
Glass&9&8&8&1&4\\
Wine&5&5&5&1&3\\
Olitos&5&5&5&6&5\\
Heart&7&8&7&4&5\\
Cleveland&11&10&10&4&7\\
Parkinson&5&6&6&3&4\\
Pima Indian Diabetes&8&8&8&2&6\\
Breast Cancer Wisconsin&7&7&7&6&7\\
Ionosphere&7&8&7&5&5\\
Sonar&5&6&6&1&5\\
Libras Movement&2&11&8&17&6\\
LSVT Voice Rehab.&5&11&7&6&7\\
Urban Land Cover&7&9&8&12&7\\
Arrhythmia&7&10&13&26&8\\
Molecular Biology&-&13&12&3&9\\
COIL 2000&29&46&33&5&8\\
CNAE-9&90&459&547&13&281\\
Madelon&-&-&-&6&7\\
MicroMass&33&168&142&24&141\\
Arcene&6&-&-&6&11\\
\hline
\end{tabular}
\label{selF}
\end{table}

Nine classifiers -- namely PART, JRip, Naive Bayes, Bayes Net, J48, BFTree, FT, NBTree and RBFNetwork -- have been chosen from different classifiers categories to classify instances of each dataset after the feature selection process. These classifiers have been implemented in Weka, a machine learning package that is ready to use \cite{Hall}. For all classifiers and the feature selection methods, 10-fold cross validation (10CV) has been conducted to calculate their performance. The mean as well as standard deviation (STD), and the best value of the nine classifiers' results over each dataset are presented in Table \ref{acc}. The best of the mean classification accuracies are boldfaced and superscripted. The last row shows the mean of the classification accuracies' mean, the STD, and the best in which the B-SFLA gains 1.22\%, 2.16\%, 2.33\%, 7.87\% higher mean classification accuracies compared to L-FRFS, GA, PSO, and GBFS, respectively. The B-SFLA outperforms other methods not only by decreasing the model size, but also by improving classification accuracy of the resulting models. Referring to the number of selected features in Table \ref{selF} and the classification accuracies in Table \ref{acc}, the GBFS has selected the least number of features and obtained the smallest classification accuracy, which is worse when compared to the unreduced datasets and to the other methods. 

\begin{table}
\caption{Mean, standard deviation, and best of classification accuracies (\%)}
\centering
\setlength{\tabcolsep}{2pt}
\begin{adjustbox}{angle=90}
\begin{tabular}{c c c | c c | c c | c c | c c}
\hline
Datasets & L-FRFS & Best & GA & Best & PSO & Best & GBFS & Best & B-SFLA & Best\\
\hline
Breast Tissue&\bf{66.46 $\pm$ 3.69}\textsuperscript{*}&70.75&\bf{66.46 $\pm$ 3.69}\textsuperscript{*}&70.75&66.46 $\pm$ 3.69&70.75&56.92 $\pm$ 4.42&61.32&65.09 $\pm$ 5.70&75.47\\
Lung Cancer&58.85 $\pm$ 12.48&77.78&41.56 $\pm$ 5.48&48.15&53.24 $\pm$ 11.53&70.37&37.04 $\pm$ 0.00&37.04&\bf{62.96 $\pm$ 12.28}\textsuperscript{*}&77.78\\
Glass&\bf{67.29 $\pm$ 7.62}\textsuperscript{*}&74.77&64.75 $\pm$ 7.76&71.96&64.75 $\pm$ 7.76&71.96&50.05 $\pm$ 5.50&54.67&65.32 $\pm$ 6.50&71.03\\
Wine&\bf{95.63 $\pm$ 2.92}\textsuperscript{*}&99.44&92.38 $\pm$ 2.23&95.51&92.38 $\pm$ 2.23&95.51&66.67 $\pm$ 1.61&68.54&93.57 $\pm$ 1.97&96.07\\
Olitos&66.39 $\pm$ 5.50&73.33&63.89 $\pm$ 3.17&68.33&65.09 $\pm$ 3.29&70.00&\bf{70.93 $\pm$ 4.24}\textsuperscript{*}&75.83&69.17 $\pm$ 4.06&77.50\\
Heart&78.48 $\pm$ 1.88&80.37&78.72 $\pm$ 1.55&80.74&\bf{79.55 $\pm$ 3.77}\textsuperscript{*}&84.07&75.93 $\pm$ 2.10&78.89&78.85 $\pm$ 1.94&81.85\\
Cleveland&49.76 $\pm$ 5.58&54.88&50.73 $\pm$ 4.87&54.88&50.73 $\pm$ 4.87&54.88&\bf{52.64 $\pm$ 2.84}\textsuperscript{*}&54.88&50.88 $\pm$ 4.11&54.88\\
Parkinson&85.07 $\pm$ 4.18&90.77&85.19 $\pm$ 3.20&90.26&83.36 $\pm$ 3.75&89.23&85.75 $\pm$ 3.31&90.26&\bf{86.50 $\pm$ 3.61}\textsuperscript{*}&89.74\\
Pima Indian Diabetes&75.00 $\pm$ 1.23&77.34&75.00 $\pm$ 1.23&77.34&75.00 $\pm$ 1.23&77.34&64.76 $\pm$ 0.95&66.15&\bf{75.35 $\pm$ 1.28}\textsuperscript{*}&76.69\\
Breast Cancer Wisconsin&96.23 $\pm$ 1.04&97.51&\bf{96.40 $\pm$ 0.54}\textsuperscript{*}&97.36&96.13 $\pm$ 0.60&96.93&95.15 $\pm$ 0.85&96.05&96.03 $\pm$ 0.92&97.36\\
Ionosphere&\bf{91.39 $\pm$ 1.04}\textsuperscript{*}&93.16&89.78 $\pm$ 1.22&92.02&89.49 $\pm$ 2.54&94.02&89.21 $\pm$ 1.40&91.74&89.65 $\pm$ 1.43&91.74\\
Sonar&69.82 $\pm$ 2.60&72.60&69.76 $\pm$ 2.29&73.08&64.26 $\pm$ 2.54&68.75&55.29 $\pm$ 3.69&61.06&\bf{74.09 $\pm$ 3.45}\textsuperscript{*}&78.85\\
Libras Movement&21.76 $\pm$ 7.45&28.61&58.14 $\pm$ 10.11&73.94&57.73 $\pm$ 7.68&67.99&\bf{61.36 $\pm$ 9.73}\textsuperscript{*}&74.17&53.43 $\pm$ 8.00&65.56\\
LSVT Voice Rehab.&79.45 $\pm$ 4.39&86.51&67.99 $\pm$ 8.10&76.98&74.52 $\pm$ 4.85&84.13&74.69 $\pm$ 10.17&80.95&\bf{79.62 $\pm$ 5.66}\textsuperscript{*}&85.71\\
Urban Land Cover&\bf{80.07 $\pm$ 2.68}\textsuperscript{*}&84.89&63.18 $\pm$ 2.87&74.37&56.50 $\pm$ 1.80&71.26&51.84 $\pm$ 1.73&83.70&77.66 $\pm$ 2.29&81.04\\
Arrhythmia&53.74 $\pm$ 3.10&57.52&53.60 $\pm$ 3.69&57.74&52.21 $\pm$ 4.52&56.42&\bf{69.05 $\pm$ 2.59}\textsuperscript{*}&74.34&60.50 $\pm$ 4.11&64.60\\
Molecular Biology&-&-&63.18 $\pm$ 1.66&65.27&56.50 $\pm$ 1.45&59.00&51.84 $\pm$ 0.17&52.19&\bf{80.12 $\pm$ 1.20}\textsuperscript{*}&81.38\\
COIL 2000&92.79 $\pm$ 2.01&94.02&92.42 $\pm$ 2.56&94.02&92.51 $\pm$ 2.40&94.02&93.97 $\pm$ 0.07&94.04&\bf{93.98 $\pm$ 0.06}\textsuperscript{*}&94.02\\
CNAE-9&\bf{88.78 $\pm$ 1.94}\textsuperscript{*}&91.57&85.77 $\pm$ 2.71&90.65&88.04 $\pm$ 3.46&92.59&53.60 $\pm$ 4.37&55.74&74.47 $\pm$ 2.32&77.96\\
Madelon&-&-&-&-&-&-&49.58 $\pm$ 0.72&50.80&\bf{54.66 $\pm$ 0.68}\textsuperscript{*}&55.40\\
MicroMass&57.40 $\pm$ 5.16&66.90&\bf{68.42 $\pm$ 5.44}\textsuperscript{*}&80.04&65.27 $\pm$ 4.10&74.78&63.07 $\pm$ 3.27&67.08&64.93 $\pm$ 4.02&73.20\\
Arcene&71.56 $\pm$ 3.00&77.00&-&-&-&-&\bf{74.94 $\pm$ 4.45}\textsuperscript{*}&81.00&70.78 $\pm$ 5.37&78.50\\
\hline
Mean&72.30 $\pm$ 3.97&77.49&71.36 $\pm$ 3.72&76.67&71.19 $\pm$ 3.90&77.20&65.65 $\pm$ 3.10&70.47&\bf{73.52 $\pm$ 3.70}\textsuperscript{*}&\bf{78.47}\textsuperscript{*}\\
\hline
\end{tabular}
\end{adjustbox}
\label{acc}
\end{table}

Table \ref{wins} shows the number of wins in terms of the best resulting classification accuracies. The L-FRFS has achieved the best accuracies for Breast Tissue, Glass, Wine, Ionosphere, Urban Land Cover, and CNAE-9. The GA has obtained the best classification accuracies in three cases, Breast Tissue, Breast Cancer Wisconsin and MicroMass. The PSO has obtained the highest classification accuracy for Heart dataset. The GBFS has achieved the best classification accuracies for five datasets -- namely, Olitos, Cleveland, Libras Movement, Arrhythmia, and Arcene. Finally, B-SFLA has reached to the maximum number of wins for eight datasets -- namely, Lung Cancer, Parkinson, Pima Indian Diabetes, Sonar, LSVT Voice Rehab., Molecular Biology, COIL 2000, and Madelon.

\begin{table}[h]
\caption{Number of wins for each method in gaining highest classification accuracy}
\centering
\begin{tabular}{c c c c c c}\hline
Algorithm&L-FRFS&GA&PSO&GBFS&B-SFLA\\
\hline
Wins&6&3&1&5&\bf{8}\\
\hline\end{tabular}
\label{wins}
\end{table}

The results of the classification accuracies have been analyzed using a non-parametric statistical comparison in KEEL software \cite{Alcala}. Non-parametric statistical analysis has been conducted in two steps. First, the equivalence of different algorithms is examined using a $1 \times N$ Friedman test. This test compares the best algorithm by ranking it against the others. The average rankings of the algorithms over 22 datasets are presented in Table \ref{avgrank}. As shown in the table, the B-SFLA approach performs best (i.e. it has the lowest ranking).

The Friedman statistic and \emph{p}-value, which is distributed according to a chi-square distribution with four degrees of freedom, are equal to 10.981818 and 0.026769, respectively. Based on the rankings in Table \ref{avgrank}, if the null-hypothesis is rejected, that is, all the algorithms achieve the same performance, then the next step is performed. In the second step, a post-hoc procedure is used to detect whether the differences between algorithms are significant or not. The \emph{p}-values for this, obtained by applying the Li post-hoc method over the results of the Friedman procedure, are presented in Table \ref{post}. Li's procedure is a powerful post-hoc method and rejects those hypotheses that have a \emph{p}-value $\le 0.043057$. Referring to the results of the post-hoc test presented in Table \ref{post}, it is clear that the B-SFLA approach performs significantly better than the PSO, GA, GBFS methods, and slightly better than L-FRFS in terms of resulting classification accuracies.

\begin{table}[h]
\caption{Average Rankings of the algorithms (Friedman) based on resulting classification accuracies}
\centering
\begin{tabular}{c c c c c c}\hline
Algorithm&L-FRFS&GA&PSO&GBFS&B-SFLA\\\hline
Ranking&2.7727&3.1818&3.4091&3.5000&\bf{2.1364}\\\hline
\end{tabular}
\label{avgrank}
\end{table}

\begin{table}[!h]
\caption{Post Hoc comparison Table for $\alpha=0.05$ (FRIEDMAN)}
\centering
\begin{tabular}{ccccc}\hline
$i$&algorithm&$z=(R_0 - R_i)/SE$&$p$&Li\\
\hline4&GBFS&2.860388&0.004231&0.043057\\3&PSO&2.669695&0.007592&0.043057\\2&GA&2.192964&0.02831&0.043057\\1&L-FRFS&1.334848&0.181926&0.05\\\hline
\end{tabular}
\label{post}
\end{table}

To compare the required computation time of the B-SFLA with the other evolutionary algorithms presented in this paper (i.e. GA and PSO), we have provided a bar graph as depicted in Figure \ref{comparison}. 

\definecolor{bblue}{HTML}{4F81BD}
\definecolor{rred}{HTML}{C0504D}
\definecolor{ggreen}{HTML}{9BBB59}
\definecolor{ppurple}{HTML}{9F4C7C}

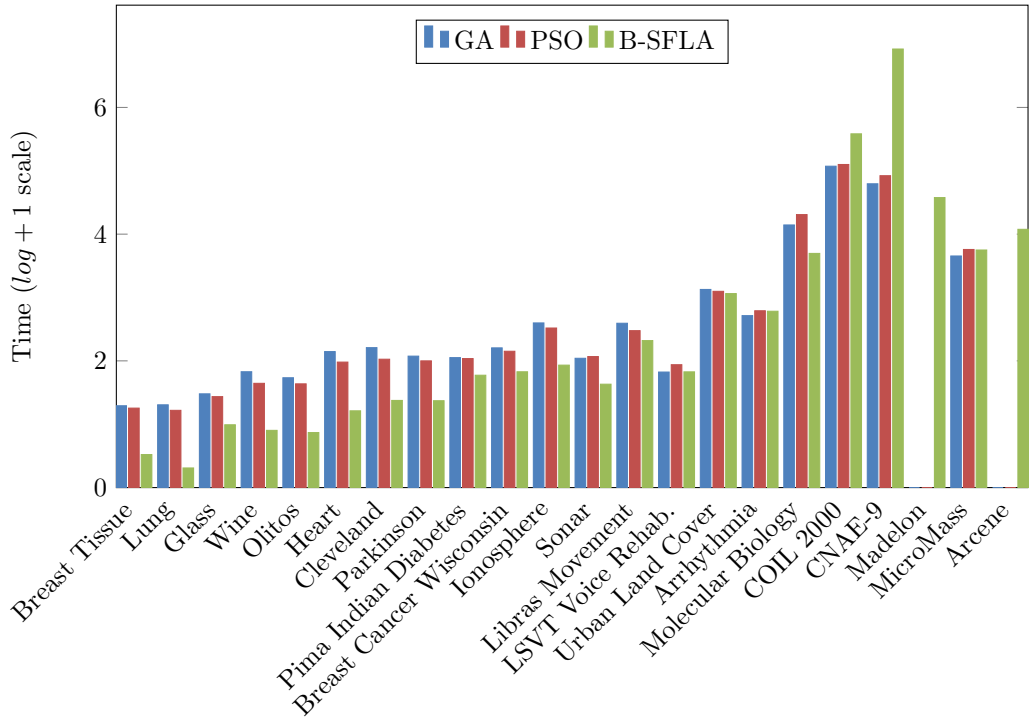
\begin{figure}[!h]
\begin{tikzpicture}
\hspace{-15mm}
    \begin{axis}[
        width  = 1.13*\textwidth,
        height = 8cm,
        major x tick style = transparent,
        ybar=2*\pgflinewidth,
        bar width=4pt,
        ylabel = {Time ($log + 1$ scale)},
        symbolic x coords={Breast Tissue,Lung,Glass,Wine,Olitos,Heart,Cleveland,Parkinson,Pima Indian Diabetes,Breast Cancer Wisconsin,Ionosphere,Sonar,Libras Movement,LSVT Voice Rehab.,Urban Land Cover,Arrhythmia,Molecular Biology,COIL 2000,CNAE-9,Madelon,MicroMass,Arcene},
        xtick = data,
        scaled y ticks = false,
        enlarge x limits=.02,
        x tick label style={rotate=45,anchor=east},
        ymin=0,
        legend style={at={(0.5,0.97)},
		anchor=north,legend columns=-1},
    ]
    
      \addplot[style={bblue,fill=bblue,mark=none}] coordinates {(Breast Tissue,1.2913688505)(Lung,1.3062105082)(Glass,1.4794313372)(Wine,1.8281441073)(Olitos,1.7333577879)(Heart,2.1462521021)(Cleveland,2.2084413564)(Parkinson,2.0745238879)(Pima Indian Diabetes,2.0510753077)(Breast Cancer Wisconsin,2.2049877042)(Ionosphere,2.598615441)(Sonar,2.0406023401)(Libras Movement,2.5935961666)(LSVT Voice Rehab.,1.8234742292)(Urban Land Cover,3.1270918342)(Arrhythmia,2.7135241429)(Molecular Biology,4.1449190157)(COIL 2000,5.0722132697)(CNAE-9,4.7945906479)(Madelon,0)(MicroMass,3.6552575612)(Arcene,0)};
            
       \addplot[style={rred,fill=rred,mark=none}] coordinates {(Breast Tissue,1.2543063323)(Lung,1.2190603324)(Glass,1.4352071032)(Wine,1.6454222693)(Olitos,1.6370892735)(Heart,1.9804578923)(Cleveland,2.0259609099)(Parkinson,1.9998262475)(Pima Indian Diabetes,2.0359897569)(Breast Cancer Wisconsin,2.1515537045)(Ionosphere,2.5183033218)(Sonar,2.0677402029)(Libras Movement,2.47665776)(LSVT Voice Rehab.,1.9391197176)(Urban Land Cover,3.0974100323)(Arrhythmia,2.7905384872)(Molecular Biology,4.3084175609)(COIL 2000,5.099225671)(CNAE-9,4.922235368)(Madelon,0)(MicroMass,3.7579514489)(Arcene,0)};
       
       \addplot[style={ggreen,fill=ggreen,mark=none}] coordinates {(Breast Tissue,0.5211380837)(Lung,0.3096301674)(Glass,0.9912260757)(Wine,0.903089987)(Olitos,0.8680563618)(Heart,1.2113875529)(Cleveland,1.3754807146)(Parkinson,1.3708830168)(Pima Indian Diabetes,1.773347542)(Breast Cancer Wisconsin,1.8275631113)(Ionosphere,1.9325752235)(Sonar,1.6317480744)(Libras Movement,2.3197512934)(LSVT Voice Rehab.,1.8263987822)(Urban Land Cover,3.0617162932)(Arrhythmia,2.7831600711)(Molecular Biology,3.6958667669)(COIL 2000,5.5830626813)(CNAE-9,6.9217436842)(Madelon,4.5768272117)(MicroMass,3.7496845372)(Arcene,4.0761268499)};

        \legend{GA, PSO, B-SFLA}
    \end{axis}
\end{tikzpicture}
\caption{Comparing computation time of three evolutionary algorithms GA, PSO and \mbox{B-SFLA}}
\label{comparison}
\end{figure}

All methods have been run on a machine with the following specifications:

\begin{itemize}
\item OS
	\begin{itemize}
		\item ubuntu 14.04 LTS
	\end{itemize}
\item Hardware
	\begin{itemize}
		\item CPU: Intel$^\text{\textregistered}$Core\texttrademark i5-4570 CPU @ 3.20GHz $\times$ 4
		\item RAM: 24 GB
	\end{itemize}
\item Software
	\begin{itemize}
		\item gcc version 4.8.4
		\item MATLAB 9.2.0.556344 (R2017a)
		\item Weka 3.6.11
		\item Java\texttrademark SE Runtime Environment (build 1.8.0\_151-b12)
	\end{itemize}
\end{itemize}

It is worth noting that computation time is highly depended on implementation methodology, optimizations and employed programming language. In this experiment, Java implementation of GA and PSO in Weka, multi-threaded C++ version of B-SFLA have been assessed. Based on the resulting computation times, B-SFLA is the fastest algorithm compared to GA and PSO except for LSVT Voice Rehab., Arrhythmia and MicroMass marginally, and for \mbox{COIL 2000} and CNAE-9 noticeably.

It is concluded that the B-SFLA is the most suitable search algorithm for FS based on the fuzzy-rough sets approach in terms of the resulting classification accuracy. Note that the B-SFLA divides the population into subpopulations, and thereby the diversity in the population is preserved. Such a swarm algorithm is very suitable for multi-modal optimization problems that have several optima instead of just one global optimum \cite{Wong}. The feature selection based on fuzzy-rough set is an example of such problems. The main intention in the L-FRFS is to obtain the minimal reducts; there exist several minimal-reducts for a given information system that are feature subsets with the minimal possible size and maximal possible FRDD. In a single run, GA and PSO generally produce one minimal reduct for a given problem as the final solution of the L-FRFS. However, the B-SFLA returns almost all of the minimal reducts in a single run in its final population. On the other hand, the B-SFLA apparently demonstrates its suitability for solving multi-modal problems since it inherently divides the population of frogs into different subpopulations. Therefore, each of these subpopulations is able to explore and exploit one of the several existing optima in the search space. This property of the B-SFLA makes it different from the other algorithms such as GA and PSO. 

\section{Conclusion and Future Work} \label{con}
In this paper, a new version of the B-SFLA has been combined with the FRDD. Additionally, the performances of L-FRFS, two well-known evolutionary algorithms, the GBFS and the B-SFLA have been compared statistically by a non-parametric statistical test. By considering the results of the statistical analysis, the B-SFLA approach significantly outperforms the PSO, GA, and GBFS methods, and is slightly better than L-FRFS in terms of resulting classification accuracy. Feature selection via fuzzy-rough theory is a multi-modal problem, i.e. there are some feature subsets with the same size and FRDD. In this sense, the B-SFLA is a suitable search algorithm for such problems, since it divides the population into subpopulations (called memeplexes), and by preserving the diversity, it returns multiple minimal reducts rather than returning just a single one. This means that several minimal reducts (i.e. the feature subsets with the minimum cardinality and maximum FRDDs) have been produced in a single run. This characteristic is an additional advantage of the B-SFLA over the PSO and GA algorithms. We are planning to apply our proposed method on local datasets, such as existing health data from Newfoundland and Labrador Centre for Health Information (NLCHI), and global ones in Canada, such as data from Statistics Canada. Also, we are aiming to improve time and space complexity of the B-SFLA to target big data, and perform comprehensive examinations and comparisons with the newly introduced feature selection methods.

\bibliographystyle{plain}
\bibliography{mybib}

\end{document}